\documentclass{article}

\usepackage[english]{babel}
\usepackage{booktabs}

\usepackage[letterpaper,top=2cm,bottom=2cm,left=3cm,right=3cm,marginparwidth=1.75cm]{geometry}

\usepackage{amsmath}
\usepackage{graphicx}
\usepackage{array}
\usepackage{booktabs}
\usepackage{tikz}
\usepackage{arydshln}
\usepackage{pgfplots}
\usepackage[colorlinks=true, allcolors=blue]{hyperref}

\title{SleepCoT: A Lightweight Personalized Sleep Health Model via Chain-of-Thought Distillation}
\author{huimin zheng,xiaofeng xing, xiangmin xu}
\pgfplotsset{compat=1.18}
\begin{document}
\maketitle

\begin{abstract}
We present a novel approach to personalized sleep health management using few-shot Chain-of-Thought (CoT) distillation, enabling small-scale language models (\(<\) 2B parameters) to rival the performance of large language models (LLMs) in specialized health domains. Our method simultaneously distills problem-solving strategies, long-tail expert knowledge, and personalized recommendation capabilities from larger models into more efficient, compact models. Unlike existing systems, our approach offers three key functionalities: generating personalized sleep health recommendations, supporting user-specific follow-up inquiries, and providing responses to domain-specific knowledge questions. We focus on sleep health due to its measurability via wearable devices and its impact on overall well-being. Our experimental setup, involving GPT-4o for data synthesis, Qwen-max for instruction set creation, and Qwen2.5 1.5B for model distillation, demonstrates significant improvements over baseline small-scale models in penalization, reasoning, and knowledge application. Experiments using 100 simulated sleep reports and 1,000 domain-specific questions shows our model achieves comparable performance to larger models while maintaining efficiency for real-world deployment. This research not only advances AI-driven health management but also provides a novel approach to leveraging LLM capabilities in resource-constrained environments, potentially enhancing the accessibility of personalized healthcare solutions.
\end{abstract}

\section{Introduction}

The rapid proliferation of wearable devices has ushered in a new era of personal health data collection. These devices, ranging from smartwatches to fitness trackers, continuously gather a wealth of physiological data that serve as external indicators of an individual's health status. Heart rate variability, sleep patterns, physical activity levels, and other metrics provide a comprehensive picture of one's well-being, offering unprecedented insights into personal health trends and potential issues.The volume and variety of data generated by these devices present both opportunities and challenges in the field of personalized healthcare. On one hand, this data deluge allows for a more nuanced and individualized understanding of health patterns. By analyzing these extensive datasets, it becomes possible to discern subtle changes in health status, potentially identifying early warning signs of various conditions or tracking the effectiveness of lifestyle changes.

The rapid evolution of large language models (LLMs), such as GPT-4o, Claude 3.5 Sonnet, and Qwen-max\cite{qwen}, has significantly advanced the handling of personalized health management and domain-specific knowledge applications. These state-of-the-art models are highly capable of generating personalized lifestyle recommendations based on physiological signals from wearable devices, such as heart rate variability and sleep patterns. By leveraging complex reasoning and expert knowledge, LLMs can offer tailored advice, enhancing users' health management in various contexts. However, despite their outstanding performance, these models face significant limitations in real-world applications, particularly in resource-constrained environments.

Firstly, the sheer size of these models poses a considerable challenge. With billions of parameters, large models demand substantial computational resources, making them impractical for deployment on edge devices like smartphones or wearable products. This restricts their usability in everyday scenarios, where on-device, real-time interactions are crucial for effective personalized health management. Furthermore, the high computational complexity associated with these models leads to increased energy consumption and high hardware costs, which hinder the widespread adoption of such advanced AI-driven solutions for individual users.Secondly, the latency associated with large models is a critical bottleneck. While these models perform well in cloud environments with abundant computational resources, their response times often fall short of user expectations due to their massive scale. In scenarios involving personal queries and health consultations, such delays can significantly degrade the user experience, failing to meet the immediate feedback requirements essential for interactive applications. Users seeking real-time insights and guidance cannot afford the lag induced by these oversized models, ultimately compromising the effectiveness of AI-driven health management.

Conversely, smaller models (with fewer than 2B parameters) can be efficiently deployed on edge devices, meeting latency requirements and allowing real-time, on-device interactions. However, their reduced size comes at the cost of diminished performance in several critical areas: question-answering reasoning abilities, handling of long-tail knowledge, and generating highly personalized recommendations. Compared to their larger counterparts, small models often struggle to match the complex reasoning capabilities required for nuanced, domain-specific guidance. This gap significantly limits their effectiveness in personalized health management, where accuracy, context-awareness, and expert-level knowledge application are crucial.

Addressing these challenges, we propose SleepCoT, a lightweight personalized sleep health model that leverages Chain-of-Thought (CoT) distillation. Our approach distills the reasoning abilities, expert knowledge, and recommendation strategies from large LLMs into compact models, thereby bridging the gap between performance and deployability. SleepCoT captures the essence of large-scale model capabilities within a smaller footprint, enabling efficient, on-device interactions without compromising the quality of personalized health management.

Our approach centers on four key areas: data synthesis, long-tail knowledge processing, personalized question-answering, and personalized recommendation generation, each designed to overcome specific limitations of small models in the context of health management.Firstly, data synthesis plays a crucial role in our methodology. By utilizing GPT-4o, a state-of-the-art large model, we generate synthetic datasets that mimic real-world sleep health scenarios, providing a diverse and robust training ground for the distillation process. This synthesized data captures a wide range of sleep patterns and health conditions, allowing our smaller model to learn complex associations that are otherwise difficult to derive from limited, real-world data. By simulating varied and challenging conditions, we enhance the small model’s ability to generalize across different user profiles, improving its reliability and effectiveness in providing personalized health insights.Secondly, long-tail knowledge processing and personalized question-answering are areas where small models often struggle due to their limited parameter capacity. Large models are proficient in handling rare, domain-specific queries and providing nuanced answers due to their extensive training on vast and diverse datasets. Through CoT distillation, SleepCoT captures and retains these critical knowledge components from larger models, significantly enhancing the small model’s ability to respond accurately to personalized, domain-specific questions. This approach enables SleepCoT to offer expert-level insights and address individualized inquiries that go beyond common health advice, effectively bridging the knowledge gap between small and large models.Finally, personalized recommendation generation is crucial for providing actionable health advice tailored to individual user needs. Large models are adept at generating personalized suggestions based on intricate data inputs such as sleep metrics, activity levels, and lifestyle factors. However, their deployment constraints hinder their practical use for real-time, interactive health management. SleepCoT addresses this by distilling the recommendation strategies of large models into a compact and efficient format, allowing the small model to deliver meaningful and personalized advice in real-time without the need for extensive computational resources. This ensures that users receive high-quality, context-aware recommendations instantly, enhancing their overall health management experience.

In summary, SleepCoT tackles the critical limitations of small-scale personalized health management models by integrating advanced data synthesis, enhancing long-tail knowledge and personalized question-answering abilities, and refining recommendation generation capabilities. This research not only advances the field of AI-driven health management but also offers a scalable, practical solution for deploying sophisticated, personalized healthcare tools in resource-constrained environments, significantly improving the accessibility and effectiveness of personalized health solutions.

\section{Related Work}

\subsection{LLM for Health applications}

Large language models exhibit the remarkable capability to provide proficient responses to free-text queries, demonstrating a nuanced understanding of professional medical knowledge \cite{javaid2023chatgpt,nazi2024large,saab2024capabilities}. Through the provision of personalised recommendations \cite{ali2023using,sudarshan2024agentic,kiser2024performance}, customised treatment strategies, and continual monitoring of patients’ advancements throughout their medical journeys, LLMs offer the promise of revolutionizing healthcare delivery. LLMs are also widely used in healthcare-related question-answering tasks\cite{yu2024enhancing,reichenpfader2024large}, where they demonstrate the ability to understand complex medical queries, provide evidence-based answers\cite{schimanski2024towards}, and offer personalized insights\cite{oduro2023enabling,neveditsin2024clinical}, making them valuable tools in clinical decision support\cite{singh2024question}, patient education\cite{yagnik2024medlm}, and health management scenarios\cite{wang2024jmlr}. The two works most closely related to ours are: one is PH-LLM\cite{cosentino2024towards}, a fine-tuned model for contextualizing physiological data and producing personalized insights, the other is PhysioLLM\cite{fang2024physiollm},  an interactive system that
leverages large language models (LLMs) to provide personalized health understanding and exploration by integrating physiological data from wearables with contextual information. Unlike their direct use of large models like GPT-3.5 or Gemini, our work focuses primarily on smaller large language models (LLMs) with parameters less than 2 billion. This approach aims to balance the performance benefits of LLMs with the practical constraints of deployment, such as computational efficiency, lower latency, and suitability for real-time, on-device applications. By concentrating on smaller models, our work addresses the challenges of making advanced AI-driven personalized health management accessible and deployable in everyday scenarios, where larger models are often impractical due to their resource demands.

\subsection{LLM with wearable data}
Personal health data, often derived from personal devices such as wearables, are distinguished by their multi-dimensional,continuous and longitudinal measurements that capture granular observations of physiology and behavior in-situ rather than in a clinical setting\cite{merrill2024transforming}. Research on LLMs based on wearable data is still in its early stages, exploring the integration of physiological data such as heart rate, sleep patterns, and physical activity into AI-driven models\cite{cosentino2024towards}. These initial studies aim to harness the potential of LLMs to provide personalized health insights and recommendations. However, as mentioned in \cite{merrill2024transforming},large-scale wearable data remains a significant barrier to the application of LLMs in personalized health scenarios. \cite{merrill2024transforming} proposes a framework for data synthesis, aiming to overcome the challenges posed by the variability, privacy concerns, and volume of wearable data. This framework seeks to generate realistic and diverse synthetic datasets that can be used to train LLMs, thereby facilitating their adaptation to personalized health applications without the need for massive real-world data collection. Derived from the aforementioned approach, but distinct in its focus, this paper specifically addresses sleep scenarios by synthesizing heart rate variability (HRV) data. Unlike prior methods, our approach generates HRV datasets that encompass a wide range of health conditions and sleep states, capturing the intricate variations in physiological responses associated with different sleep qualities. This targeted synthesis of HRV data is crucial for accurately reflecting the complex interactions between sleep and overall health, enhancing our model’s capacity to deliver personalized recommendations and insights tailored to individual sleep health management.

\subsection{LLM distillation}
Due to resource constraints and real-time requirements\cite{huang2024new}, many studies have focused on distilling large language models, which involves transferring the knowledge\cite{mcdonald2024reducing,yang2024survey}, reasoning capabilities\cite{li2024turning,kang2024knowledge}, and specialized domain expertise\cite{yuan2024llm} from these expansive models into smaller, more efficient models. \cite{shridhar2022distilling} proposes an alternative reasoning scheme, SOCRATIC COT that learns a decomposition of the original problem into a
sequence of subproblems and uses it to guide the intermediate reasoning steps. \cite{kang2024knowledge}  proposes Knowledge-Augmented Reasoning Distillation (KARD), a novel method that fine-tunes small LMs to
generate rationales obtained from LLMs with augmented knowledge retrieved from an external knowledge base. However, these methods are primarily oriented towards numerical reasoning and factual judgment, focusing on interpreting data points and deriving logical conclusions. They excel in processing quantitative information and making evidence-based decisions but often lack the depth needed to address the more subjective, nuanced aspects of personalized health management, such as interpreting subtle physiological changes or providing context-aware, personalized advice.  Many studies\cite{wei2022chain,kojima2022large,fu2023specializing} have also utilized Chain-of-Thought (CoT) methods for distillation, which are particularly well-suited for scenarios involving personalized sleep-related question-answering and recommendation generation. CoT distillation allows smaller models to inherit the complex, step-by-step reasoning capabilities of larger models, making them more adept at understanding and responding to personalized queries and generating tailored sleep health advice. This approach aligns well with the nuanced nature of sleep health management, where sequential reasoning and context-aware guidance are essential.

\section{MOTIVATION}
\begin{figure}
    \centering
    \includegraphics[width=\linewidth]{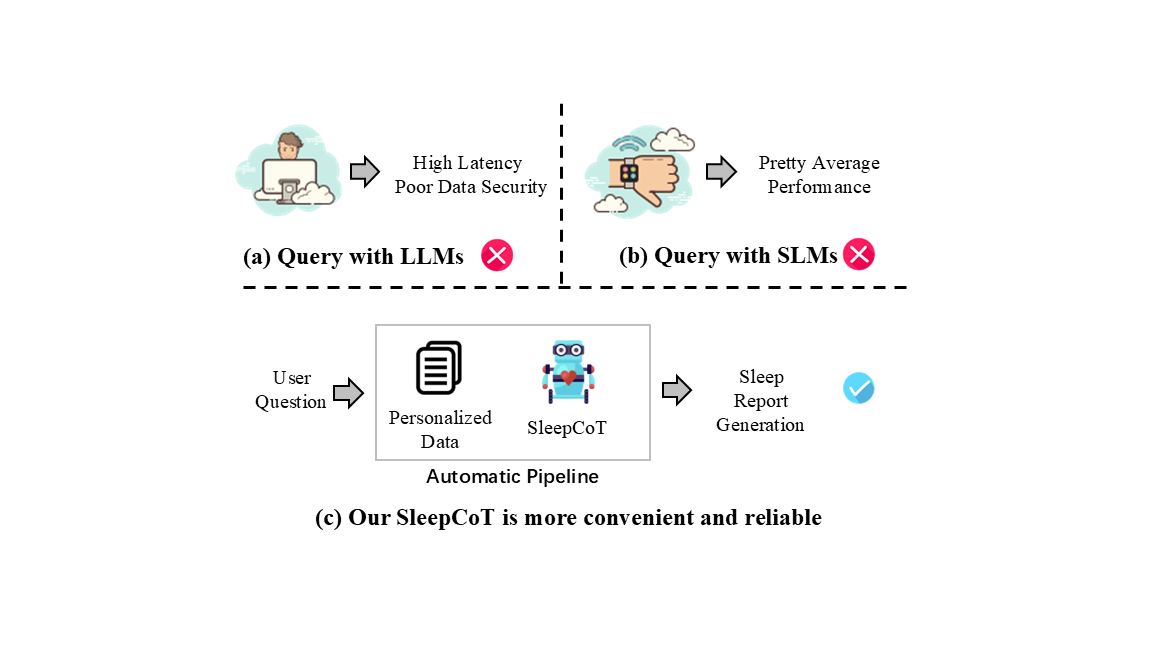}
    \caption{Motivation and Goal}
    \label{fig:motivations}
\end{figure}

As demonstrated in \cite{fang2024physiollm,cosentino2024towards}, state-of-the-art large models, such as GPT-4o and Qwen-max, already possess capabilities for personalized question-answering and data interpretation specifically tailored to the sleep health domain. These advanced models can understand complex sleep-related data and provide customized health advice directly. However, their substantial size and computational demands make them unsuitable for deployment in resource-constrained environments or applications with high real-time requirements. Simultaneously, a substantial body of research\cite{fu2023specializing,upadhayayaya2024efficient,tian2024tinyllm} has shown that in specific domains, smaller models can attain performance levels comparable to those of larger models through the process of distillation. Distillation enables these smaller models to effectively inherit the specialized knowledge, complex reasoning capabilities, and contextual understanding from their larger counterparts, thereby enhancing their efficiency and effectiveness despite having significantly fewer parameters. These findings validate the potential of using distilled smaller models in specialized applications, where they can provide high-quality outcomes while addressing practical constraints such as limited computational resources and stringent real-time requirements, as shown in \ref{fig:motivations}
This motivation drives our design goals: to create a model that not only inherits the advanced understanding and personalized response capabilities of larger models but also addresses the practical constraints of deployment. By achieving a balance between performance and deployability, our work seeks to bring advanced, AI-driven personalized sleep health management closer to everyday use, offering the potential for real-time, personalized insights directly at the user's fingertips.

\section{SleepCoT ARCHITECTURE AND IMPLEMENTATION}
\begin{figure}
    \centering
    \includegraphics[width=\linewidth]{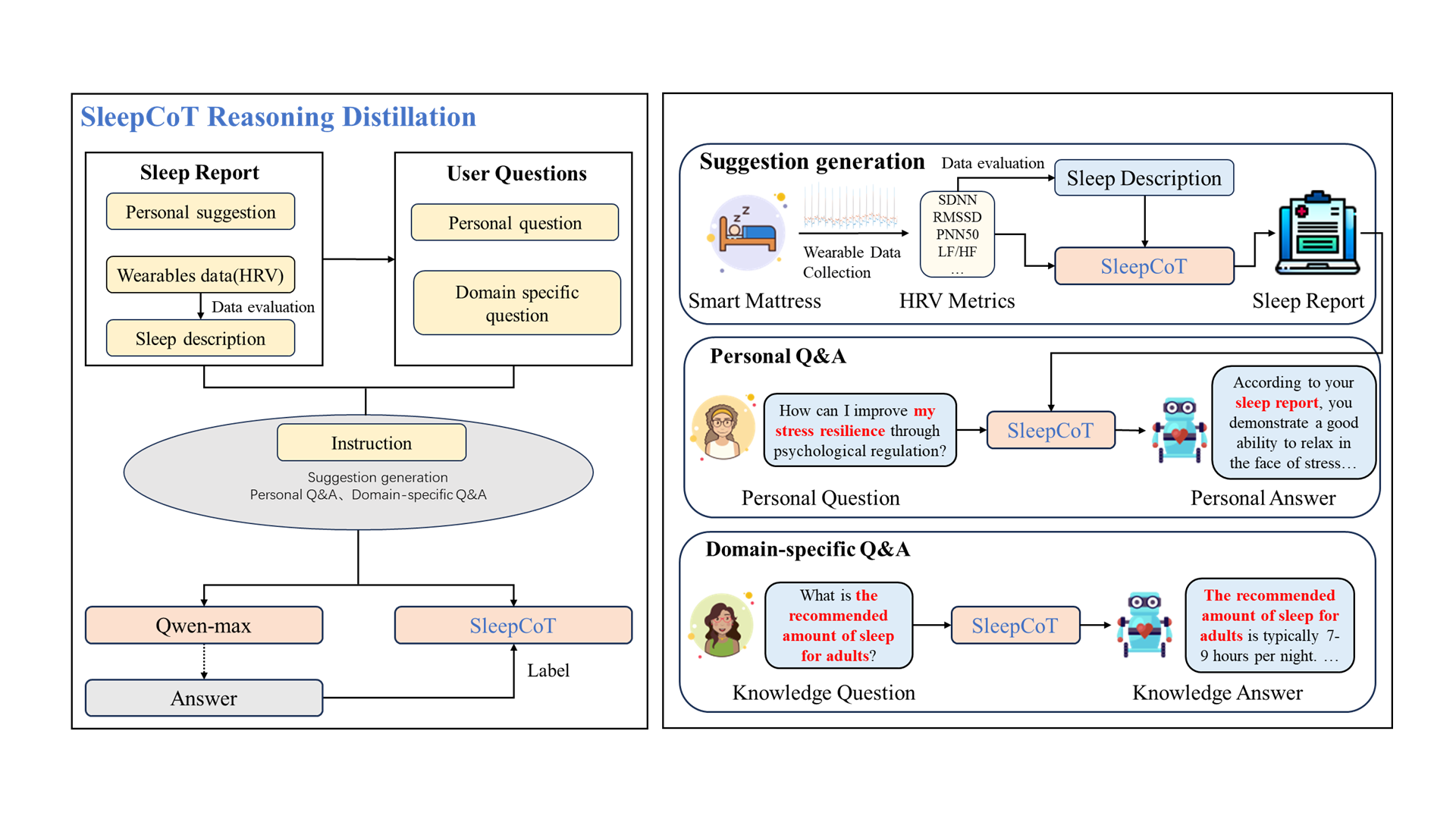}
    \caption{ARCHITECTURE of SleepCoT}
    \label{fig:frameworks}
\end{figure}

Figure \ref{fig:frameworks} illustrates the overall framework of SleepCoT, which consists of three main components: data acquisition, personalized recommendation generation, and user question-answering (including both personalized and domain-specific knowledge inquiries).

\subsection{Data acquisition}
In real-world scenarios, wearable devices such as smartwatches or sensor-embedded mattresses can capture electrocardiogram (ECG) signals during the user's sleep. From these signals, various heart rate variability (HRV) parameters can be derived, such as SDNN (Standard Deviation of NN intervals) and RMSSD (Root Mean Square of Successive Differences). These HRV metrics provide valuable insights into the autonomic nervous system’s activity and overall sleep quality, serving as critical indicators for personalized sleep health analysis and recommendation generation.  \cite{cosentino2024towards} has demonstrated the significant potential of Large Language Models (LLMs) in synthesizing wearable data. This highlights their capability to generate realistic and diverse datasets, which can be effectively utilized for training and evaluating models in personalized health applications.
 
Figure illustrates the overall framework of data synthesis. The framework outlines the process of generating synthetic datasets, starting from the collection and processing of instance data, followed by the application of physiological parameter constraints. This approach ensures that the synthesized data closely mimics real-world conditions. The framework also includes steps for generating personalized recommendations and potential user questions based on the synthesized sleep reports, thereby creating a comprehensive and realistic dataset for model training and evaluation. Using GPT-4o, we synthesized 100 samples derived from real-world examples(\(i\)) and predefined physiological rules(\(R\)). Each sample consists of HRV parameters collected over six consecutive nights, including key metrics such as SDNN and RMSSD, alongside average sleep duration and sleep staging data, such as time spent in light, deep, and REM sleep stages. To assess users' cardiac health, stress resilience, and other related conditions, we employed predefined algorithms analogous to those described in prior studies\cite{kinnunen2020feasible,vigo2019heart,taralov2015heart}. These algorithms facilitated the generation of comprehensive sleep state assessments, which were subsequently integrated with the corresponding sample data. Leveraging GPT-4o, personalized recommendations were then generated based on these combined assessments, providing tailored guidance specific to the evaluated sleep profiles.  The above process can be represented by the following equations:

\[wearable\_data = GPT-4o(Pr1(I+R))\]

\[sleep\_description = D\_A(wearable_data)\]

\[personal\_suggestion = GPT-4o(Pr2(wearable+sleep\_description))\]

where \(D\_A\) represents the predefined evaluation algorithm, \(Pr\) represents the predefined prompt template, while Table \ref{tab:template1} and Table \ref{tab:template2} is the example prompt of \(Pr1\) and \(Pr2\).  \(I\) represents a sample obtained from wearable data.

\begin{table}[htbp]
\centering
\begin{tabular}{@{}p{0.95\textwidth}@{}}
\toprule
\textbf{Instruction Template1:} \\
The example {I} is based on 6 days of monitoring data from the user. Please generate 100 similar data in the same format, reflecting different sleep health conditions across the population. Ensure that the parameters in the reports reflect realistic and plausible health states without presenting impossible or contradictory health conditions. Additionally, pay attention to the diversity and richness of the report content, avoiding repetition. Do not omit the HRV parameters, as they are a crucial indicator in the sleep report and need to be included. Please follow the instructions and ensure each report starts with "Sleep Quality Report:". \\
\bottomrule
\end{tabular}
\caption{\label{tab:template1}Wearable Data production}
\end{table}

\begin{table}[htbp]
\centering
\begin{tabular}{@{}p{0.95\textwidth}@{}}
\toprule
\textbf{Instruction Template2:} \\
You are a sleep expert. 
Please generate personalized recommendations based on the following sleep report.
**Sleep Quality Report:**

1. **Sleep Quality Overview**
   - During the observation period, the subject's average sleep duration was 7.7 hours, meeting the recommended 7-9 hours of sleep for adults...

2. **Cardiac Health**
   - Analysis of heart rate variability (HRV) parameters, including SDNN, RMSSD, LF/HF, and PNN50, provides an assessment of the subject's autonomic nervous system balance and cardiac health status...

3. **Stress and Stress Resilience**
   - Analysis of the LF/HF ratio and HF components shows that the subject's stress level was low during the observation period, with strong parasympathetic activity...

4. **Sleep Apnea and Sleep Interruptions**
   - The subject experienced an average of 9 sleep apnea events per night...

**Comprehensive Impact Analysis**
Overall, the subject's sleep quality is good, cardiac health is well-maintained, stress levels are low, and stress resilience is strong...

- **Stress Resilience**: Good
- **Stress Level**: Low
- **Fatigue Level**: Mild
- **Autonomic Nervous System Activity**: Good
**HRV Parameters Calculation:**

- **SDNN**: [53, 55, 54, 56, 53, 54]
  - Description: General range: 20-220. For healthy adults (24-hour recording): 141±39. For short-term recordings (5 minutes): 50±16.
  
- **RMSSD**: [66, 68, 67, 69, 70, 68]
  - Description: General range: 10-50. For healthy adults: 42±15.

- **LF/HF**: [1.2, 1.3, 1.1, 1.4, 1.2]
  - Description: For healthy adults: typically between 0.5 to 2.0. 24-hour recording...

- **PNN50**: [38.0, 40.5, 39.0, 41.0, 39.5, 40.0]
  - Description: General range: approximately 0-50. For healthy adults: typically greater than 10.
\\
\bottomrule
\end{tabular}
\caption{\label{tab:template2}Personalized suggestion production}
\end{table}

After generating the personalized recommendations, we utilized GPT-4o to create a set of likely questions based on the sleep reports, which included both physiological parameters and the personalized advice. For each of the 100 sleep reports, 150 personalized questions were generated, resulting in a total of 15,000 personalized questions. It is important to note that when generating questions using GPT-4o, the 100 user reports were sequentially input in the form of multi-turn dialogues. This method helps to minimize the generation of repetitive questions and enhances the diversity of the questions produced, ensuring a broader and more varied set of queries that better reflects real-world interactions. The prompt used in this part is shown in table \ref{tab:template3}. In addition to this, GPT-4o was also employed to generate 2,000 sleep-related knowledge questions. Furthermore, Claude-Sonnet 3.5 was used to produce an additional 100 domain-specific knowledge questions.
These synthesized samples are designed to closely replicate real-world sleep conditions, providing a realistic and diverse set of data to train and test the model's ability to generate personalized recommendations and accurately respond to sleep-related queries.

\begin{table}[htbp]
\centering
\begin{tabular}{@{}p{0.95\textwidth}@{}}
\toprule
\textbf{Instruction Template3:} \\
"You are a sleep medicine expert. You need to generate 150 questions that users are most likely to ask in their daily lives based on each sleep quality report.""These questions should mainly revolve around personalized information, such as: Is my SDNN value normal? The answers to such questions require finding relevant SDNN information from the sleep report." "For example, Will my sleep condition affect my daily performance? The answers to such questions need to find relevant information about sleep conditions and daily performance from the sleep report." "It should be noted that for questions like Is my sleep efficiency <number> normal?, it should be changed to: Is my sleep efficiency normal? That is, specific numerical values should not be included in the generated questions." "In personalized Question Answering, ensure the diversity of question content and phrasing. Please generate the questions according to the instructions, and ensure all outputs start with Question 1:, with only content related to the questions." \\
\bottomrule
\end{tabular}
\caption{\label{tab:template3}Personalized question production}
\end{table}

\subsection{Experiment Setting}
Before distillation, the teacher model Qwen-max was tested on the public dataset SleepQA\cite{bojic2023building} to validate the richness of its domain-specific expertise in the field of sleep.In the distillation experiment, we set up three distillation tasks: personalized recommendation generation, personalized question-answering, and domain-specific knowledge question-answering. These three tasks were trained jointly. In addition, experiments were conducted to investigate the impact of varying the proportions of instruction sets among the three tasks on model performance. The experiment was designed to understand how different ratios of task-specific instruction sets affect the overall effectiveness of the model. By adjusting the proportions of instruction sets for tasks such as personalized question-answering, domain-specific knowledge question-answering, and recommendation generation, the goal was to identify which task data most significantly enhances model performance. This approach allows for the optimization of instruction set design and data allocation, improving the model’s performance in real-world applications. In resource-constrained environments, these findings enable the more efficient utilization of data to enhance the model’s capabilities.

\subsubsection{Instruction Set}
The training set comprises 80 suggestion generation samples, 12000 personalized question-answering samples, and 600 knowledge-based question-answering samples. The test set includes 20 suggestion generation samples, 3000 personalized question-answering samples, and 200 knowledge-based question-answering samples. Additionally, the test set contains a separate set of 100 personalized questions generated by Claude-Sonnet 3.5. 

\subsubsection{Few shot Chain-of-Thought Prompt}
In this study, the model is guided to answer questions using a Chain-of-Thought (CoT) approach, where the reasoning process mimics human thought patterns. The CoT prompts are designed to first extract relevant information from the context before generating a response. In open-ended question-answering scenarios, questions are categorized into three types to enhance the accuracy of the answers: those where information can be directly found from the context, those that require a global summary, and those where relevant information is not available in the context. Each category is illustrated with an example. These components collectively form the few-shot In-Context Learning (ICL) setup. The main part of the Few shot Chain-of-Thought is shown in Table \ref{tab:template4}

\begin{table}[htbp]
\centering
\begin{tabular}{@{}p{0.95\textwidth}@{}}
\toprule
\textbf{Prompt Template4:} \\
"You are a sleep expert. Please answer the user's questions based on their sleep report. The main approach to answering questions should be: \textbf{First, identify relevant information from the sleep report based on the question, then respond using the specific information found.} Avoid giving general answers; instead, respond based on the specific details reflected in the user's sleep report. Below is an example: \textbf{1. For questions with clearly identifiable information}: Question: Is my weight within the normal range? First, find the relevant information: According to your sleep report, your weight is <number1>kg. Then determine whether the weight of <number1>kg is within the normal range. 2. \textbf{For questions requiring a summary of the overall report}: Question: How often should I have a heart health check-up? First, perform a global summary of the sleep report, then answer the question based on the summarized result. 3. \textbf{For questions not directly addressed in the sleep report}: Question: Do I need to change my drinking habits to improve my sleep? Combine the overall sleep report with the question, then provide an answer. " \\
\bottomrule
\end{tabular}
\caption{\label{tab:template4}Few shot Chain-of-Thought}
\end{table}

\subsubsection{model and Training Parameter setting}
Qwen-max was selected as the teacher model, and Qwen2.5.5-1.5B was chosen as the student model. The models were fine-tuned using the LoRA (Low-Rank Adaptation)\cite{hu2021lora} approach. The fine-tuning process employed the LoRA (Low-Rank Adaptation) approach with specific hyper-parameters set as follows: the learning rate was configured at 1.0e-5, the batch size was set to 1, the LoRA rank was set to 8, and the number of epochs was set to 10. These settings were chosen to ensure efficient training and effective adaptation of the student model (Qwen2.5-1.5B) and (Qwen2.5-0.5B) from the teacher model (Qwen-max) while maintaining computational efficiency. In addition, experiments were conducted directly using GPT-4o\footnote{https://chatgpt.com/}, Claude-Sonnet 3.5\footnote{https://claude.ai/new}, Baichuan4\footnote{https://platform.baichuan-ai.com/playground}, GLM-4\footnote{https://open.bigmodel.cn/console/trialcenter?modelCode=glm-4-plus}, Gemini 1.5 Pro\footnote{https://deepmind.google/technologies/gemini/pro/}, Qwen2.5-7B, and Qwen2.5-1.5B to enhance the richness of the experiments.
Finally, we tested the distilled model, SleepCoT, on the SleepQA dataset to evaluate its performance in the sleep domain and assess its ability to effectively handle sleep-related questions while retaining sufficient domain knowledge and reasoning capabilities after the distillation process.
\subsubsection{Evaluation method}
Since traditional evaluation methods such as BLEU, ROUGE, and BERTScore struggle to effectively differentiate model performance in this scenario, GPT-4o\cite{bavaresco2024llms} is employed as the evaluator. Inspired by RAGAS\cite{es2023ragas}, the models are assessed based on the following four dimensions:
\begin{itemize}
    \item [$\bullet$] \textbf{Penalization}: Evaluates how well the generated recommendations and answers are tailored to the individual user’s data and specific needs.
\end{itemize}
\begin{itemize}
    \item [$\bullet$] \textbf{Relevance}: Measures the alignment of the responses with the user's context and the specific questions asked, ensuring that the information provided is pertinent.
\end{itemize}
\begin{itemize}
    \item [$\bullet$] \textbf{Completeness}: Assesses whether the responses comprehensively cover all necessary aspects of the query, ensuring no critical details are left out.
\end{itemize}
\begin{itemize}
    \item [$\bullet$] \textbf{Accuracy}: Evaluates the correctness of the information provided, focusing on domain-specific knowledge and the validity of personalized advice.
\end{itemize}
Each dimension was scored on a scale of 1 to 5, with five levels of assessment.

\subsection{Main Results}
The results of the evaluation of the teacher model Qwen-max regarding the adequacy of its domain-specific expertise in the field of sleep are shown in table \ref{tab:table_qwenmax}. As demonstrated in Table \ref{tab:table_qwenmax}, Qwen-max's domain-specific expertise in the field of sleep is not only adequate but also performs well across various evaluation metrics. This suggests that the model has a strong understanding of sleep-related knowledge and is capable of answering questions with a high degree of accuracy. The robustness of Qwen-max in this specialized domain indicates that it can serve as a reliable teacher model for downstream tasks, such as knowledge distillation, where capturing and transferring this expertise to smaller, more efficient models is crucial. After the distillation process, the distilled model retained the same level of domain expertise, showing no decline in the quality of specialized knowledge. This validates both the potential of Qwen-max to provide accurate and comprehensive insights into sleep-related queries and the effectiveness of the distillation process in preserving essential knowledge, making the distilled model a suitable candidate for further application in sleep-related QA systems.

\begin{table}[htbp]
\centering
\begin{tabular}{l|c}
\toprule
\textbf{Extractive QA system name} & \textbf{EM} \\
\midrule
\textit{Lucene BM25\cite{lin2021pyserini} + BERT SQuAD2 QA\cite{rajpurkar2018know}} & 0.30 \\
\textit{PubMedBERT + BioBERT BioASQ\cite{bojic2023building}} & 0.24 \\
\textit{Qwen-max} & 0.94 \\
\textit{SleepCoT-1.5B} & 0.92 \\
\textit{SleepCoT-0.5B} & 0.92 \\
\bottomrule
\end{tabular}
\caption{\label{tab:table_qwenmax}Performance of QA Systems}
\end{table}

\begin{figure}
    \centering
    \includegraphics[width=0.5\linewidth]{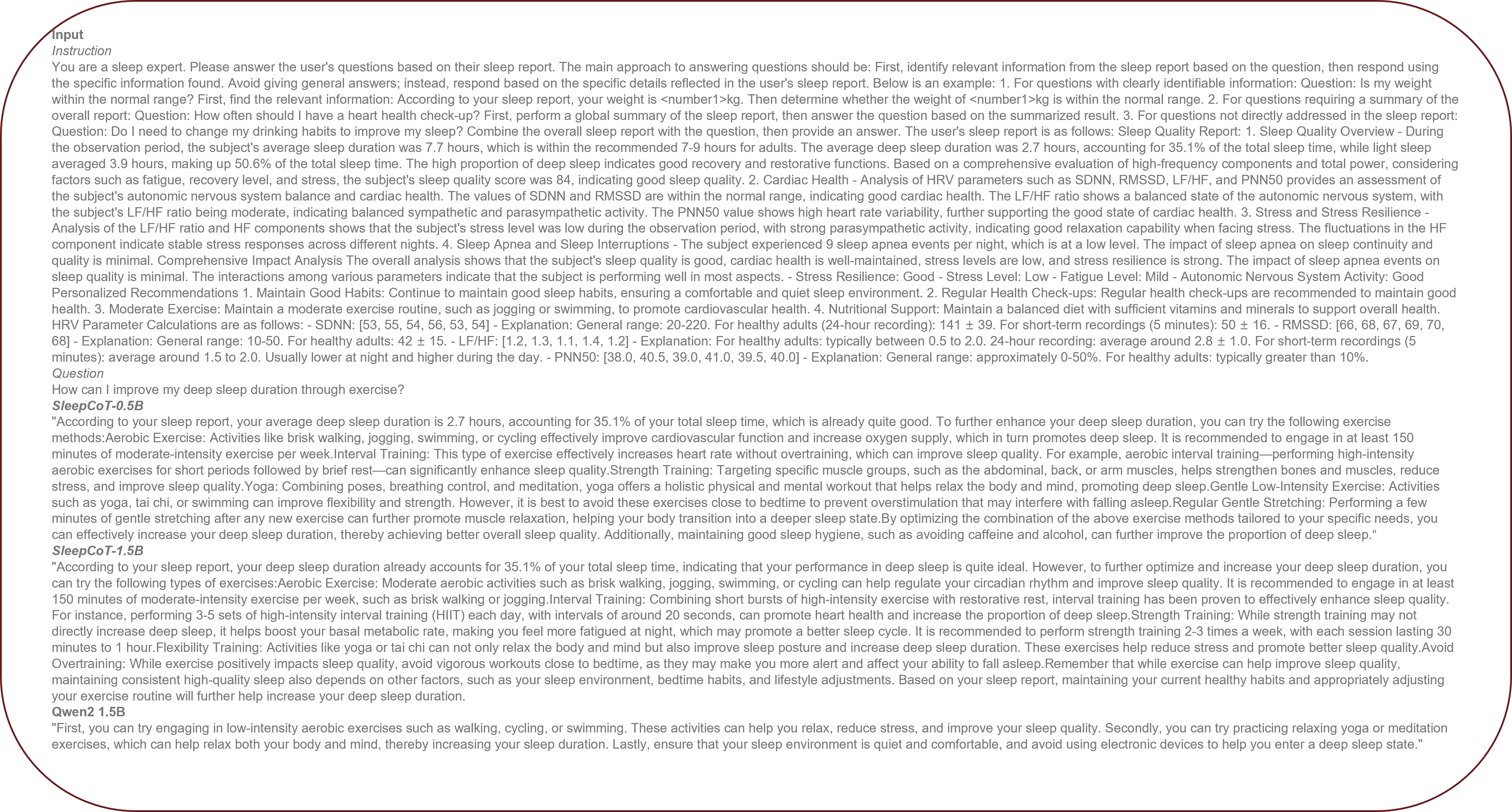}
    \caption{An example of models' response}
    \label{fig:tables}
\end{figure}

The evaluation results of SleepCoT are presented in table \ref{tab:table1}. It shows that proposed method SleepCoT improves task performance across all baselines. It can be observed that SleepCoT-1.5B performs on par with Qwen-max across all four evaluation dimensions, while SleepCoT-0.5B a slight performance gap compared to SleepCoT-1.5B. An example is shown in Figure \ref{fig:tables}. From this example, it can be observed that open-source small models like Qwen2.5-7B lack sufficient penalization in their responses. From the performance gap observed between the 0.5B and 1.5B distilled small models, it is evident that the 0.5B model lags behind in the accuracy dimension compared to the 1.5B model. This indicates that even after fine-tuning, the 0.5B model's ability to utilize long-tail knowledge is still not as strong as that of the 1.5B model.

\begin{table}[htbp]
\centering
\setlength{\tabcolsep}{6pt}
\begin{tabular}{l*{4}{c}|c}
\toprule
Models & Penalization & Relevance & Completeness & Accuracy & \textbf{Average} \\
\midrule
Qwen-max & 4.8 & 4.9 & 4.7 & 4.9 & \textbf{4.8} \\
Qwen2.5-7B & 4.0 & 4.2 & 4.3 & 4.5 & \textbf{4.25} \\
Qwen2.5-1.5B & 3.5 & 3.7 & 3.5 & 3.5 & \textbf{3.5} \\
\cdashline{1-6}
GPT-4o & \textbf{5} & \textbf{5} & \textbf{5} & \textbf{5} & \textbf{5}\\
Claude-Sonnet 3.5 & 4.6 & 4.7 & 4.5 & 4.8 & 4.7\\
Baichuan4 & 4.7 & 4.7 & 4.6 & 4.8 & 4.7\\
GLM-4 & 4.8 & 4.6 & 4.6 & 4.9 & 4.8\\
Gemini 1.5 Pro & 4.6 & 4.7 & 4.5 & 4.8 & 4.6\\
\cdashline{1-6}
sleepCoT-0.5B & 4.3 & 4.4 & 4.3 & 4.2 & \textbf{4.3} \\
SleepCoT-1.5B & 4.8 & 4.7 & 4.7 & 4.7 & \textbf{4.7} \\
\bottomrule
\end{tabular}
\caption{\label{tab:table1}Model Comparison}
\end{table}

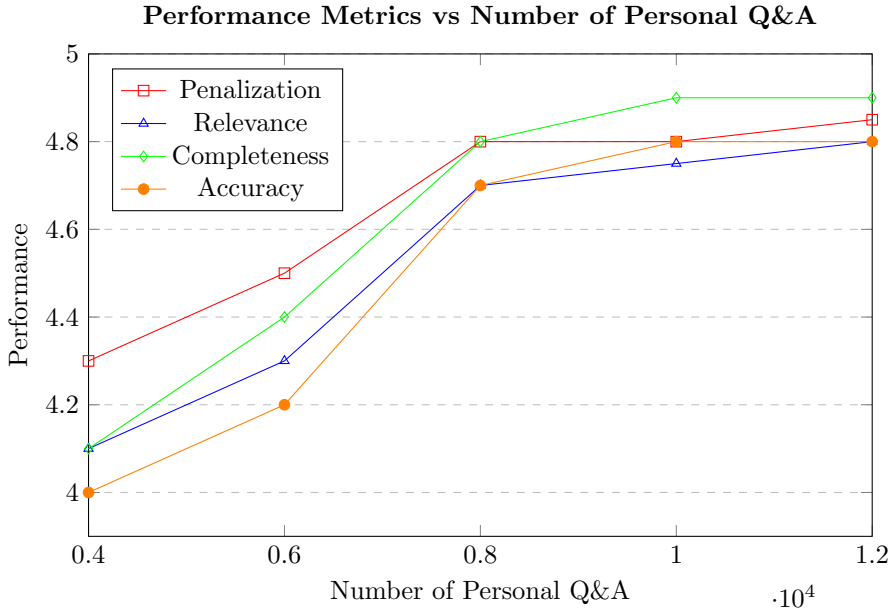
\begin{figure}[htbp]
    \centering
    \begin{tikzpicture}
    \begin{axis}[
        width=12cm,
        height=8cm,
        title={Performance Metrics vs Number of Personal Q\&A},
        title style={font=\bfseries},
        xlabel={Number of Personal Q\&A},
        ylabel={Performance},
        xmin=4000, xmax=12000,
        ymin=3.9, ymax=5,
        xtick={4000,6000,8000,10000,12000},
        ytick={4.0,4.2,4.4,4.6,4.8,5.0},
        legend pos=north west,
        ymajorgrids=true,
        grid style=dashed,
    ]

    \addplot[
        color=red,
        mark=square,
        ]
        coordinates {
        (4000,4.3)(6000,4.5)(8000,4.8)(10000,4.8)(12000,4.85)
        };
        
    \addplot[
        color=blue,
        mark=triangle,
        ]
        coordinates {
        (4000,4.1)(6000,4.3)(8000,4.7)(10000,4.75)(12000,4.8)
        };
        
    \addplot[
        color=green,
        mark=diamond,
        ]
        coordinates {
        (4000,4.1)(6000,4.4)(8000,4.8)(10000,4.9)(12000,4.9)
        };
        
    \addplot[
        color=orange,
        mark=*,
        ]
        coordinates {
        (4000,4.0)(6000,4.2)(8000,4.7)(10000,4.8)(12000,4.8)
        };
        
    \legend{Penalization,Relevance,Completeness,Accuracy}

    \end{axis}
    \end{tikzpicture}
    \caption{Performance Metrics vs Number of Personal Q\&A}
    \label{fig:performance_metrics} 
\end{figure}

The results of the experiment on adjusting data proportions are presented in Figure \ref{fig:performance_metrics}. As the number of personalized question-answering instructions increases from 4000 to 12000, there is a noticeable improvement in the scores across all four dimensions—Penalization, Relevance, Completeness, and Accuracy. 
\begin{itemize}
    \item \textbf{Penalization:} Starting at a score of 4.3 with 4000 instructions, it steadily improves, reaching 4.8 at 8000 instructions, and slightly increases to 4.85 as the data volume hits 12000. This suggests that increased personalized data significantly enhances the model's ability to tailor responses effectively.
    \item \textbf{Relevance:}  Initially scoring 4.1 at 4000 instructions, relevance improves to 4.7 by 8000 instructions and continues to rise to 4.8 at 12000. This indicates that with more data, the model becomes increasingly adept at providing responses closely aligned with the context and user queries.
    \item \textbf{Completeness:} The completeness score starts at 4.1 and shows a similar upward trend, reaching 4.8 at 8000 instructions and maintaining that level as data increases further. This trend highlights that sufficient personalized data helps the model provide more comprehensive answers.
    \item \textbf{Accuracy:}  Accuracy begins at 4.0 and gradually improves, reaching 4.8 at 8000 instructions and stabilizing there as more data is added. This plateau suggests that accuracy benefits significantly from increased data up to a point but shows diminishing returns beyond 8000 instructions.
Overall, these results depict that increasing the personalized question-answering instruction set leads to substantial performance gains across all evaluated dimensions, with diminishing returns observed as the dataset approaches 8000 entries, suggesting a performance cap.
\end{itemize}

\subsection{Ablation Study}
To demonstrate the necessity of each component in SleepCoT, we take a series of ablation study by remov-
ing the following parts:(1) examples in prompt: the instruction prompt contains only Chain of Thought.(2) examples and CoT: the instruction prompt contains nothing but the query.
The results are shown in \ref{tab:table_abaltion}, when filtering examples in prompt with same process, we find that the model performance decreases, confirming the value of examples. Further removing the CoT, Compared to the performance drop observed in the first ablation experiment, this setup resulted in a significantly greater decline in performance, indicating The CoT plays an important role in the distillation process. 

\begin{table}[h]
\centering
\begin{tabular}{l@{\hspace{0.5em}}c@{\hspace{0.5em}}c@{\hspace{0.5em}}c@{\hspace{0.5em}}c|@{\hspace{0.5em}}c}
\toprule
Models & Penalization & Relevance & Completeness & Accuracy & \textbf{Average} \\
\midrule
plain prompt & 3.4 & 3.6 & 3.4 & 3.3 & \textbf{3.5} \\
prompt with CoT & 4.4 & 4.2 & 4.1 & 4.3 & \textbf{4.3} \\
few shot CoT & 4.8 & 4.7 & 4.7 & 4.7 & \textbf{4.7} \\
\bottomrule
\end{tabular}
\caption{\label{tab:table_abaltion}Ablation experiment}
\end{table}

\section{Conclusion}
This work explores the effectiveness of few-shot chain-of-thought prompting for distilling complex reasoning abilities and domain-specific knowledge from large language models to specialized smaller models. The approach was validated in the sleep domain, where a 1.5B-sized model was shown to be easily deployable on edge devices while meeting specific demands in vertical domains, such as personalized sleep recommendations. As with other specialized fields, this method can be extended to similar domains, demonstrating its versatility and potential for broader applications.

\section{Limitations}
One limitation of this work is the potential lack of generalization to other domains, as the success in the sleep domain may not translate directly to more complex or less structured fields. Additionally, the method's effectiveness is highly dependent on the quality and diversity of the data, making it vulnerable to bias or reduced performance in cases where domain-specific data is limited. While smaller models like the 1.5B model are easier to deploy on edge devices, there remains a trade-off between efficiency and performance, particularly in more complex tasks. Moreover, few-shot learning techniques, though effective for specialized tasks, may struggle to scale to broader applications that require a more extensive knowledge base. The ability of the model to adapt in real-time to new data or evolving user needs is another potential limitation, as its personalized recommendations may lose relevance without frequent updates. Furthermore, traditional evaluation metrics such as BLEU or ROUGE may not fully capture the model's deeper reasoning capabilities, limiting the assessment of its true effectiveness. Finally, while edge deployment is feasible, optimizing the model to function effectively within the strict resource constraints of edge devices, without compromising performance, remains a challenge.

\bibliographystyle{plain}
\bibliography{references}
\section{Appendix}

\end{document}